\documentclass[journal]{IEEEtran}

\usepackage{cite}

\ifCLASSINFOpdf
  \usepackage[pdftex]{graphicx}
  \usepackage{graphics}
  \DeclareGraphicsExtensions{.pdf,.jpeg,.png}
\else
  \usepackage[dvips]{graphicx}
  \usepackage{graphics}
  \DeclareGraphicsExtensions{.eps}
\fi
\usepackage{amsmath}
\usepackage{amssymb,amsthm}
\usepackage{algorithm}
\usepackage{algorithmic}

\usepackage{threeparttable}
\usepackage{array}
\usepackage{multirow}
\usepackage{colortbl,booktabs}

\usepackage{subfigure}

\ifCLASSOPTIONcompsoc
  \usepackage[caption=false,font=normalsize,labelfont=sf,textfont=sf]{subfig}
\else
  \usepackage[caption=false,font=footnotesize]{subfig}
\fi

\ifCLASSOPTIONcaptionsoff
  \usepackage[nomarkers]{endfloat}
  \let\MYoriglatexcaption\caption
  \renewcommand{\caption}[2][\relax]{\MYoriglatexcaption[#2]{#2}}
\fi

\usepackage{url}

\usepackage{pgfplots}
\pgfplotsset{compat=1.13}

\usepackage{afterpage}
\usepackage{lscape}
\begin{document}

\title{Progressive Transfer Learning}

\author{Zhengxu~Yu,
        Dong~Shen,
        Zhongming~Jin,
        Jianqiang~Huang,
        Deng~Cai,~\IEEEmembership{Member,~IEEE} and Xian-Sheng Hua,~\IEEEmembership{Fellow,~IEEE}
%\thanks{Manuscript received ????? ?? ????; revised ????? ?? ????. This work was supported by ???????. (Corresponding author: D. Cai.)}%
\thanks{Z.~Yu, D.~Shen and D.~Cai are with the State Key Laboratory of CAD\&CG, College of Computer Science, Zhejiang University, Hangzhou, Zhejiang 310058, China (emails: yuzxfred@gmail.com; dongshen@zju.edu.cn; dengcai@gmail.com).}% 
\thanks{Z.~Jin, J.~Huang and X.-S.~Hua are with Alibaba Group, Hangzhou, Zhejiang, China (emails: zhongming.jinzm@alibaba-inc.com; jianqiang.hjq@alibaba-inc.com; xiansheng.hxs@alibaba-inc.com).}}

\ifCLASSOPTIONpeerreview
\markboth{IEEE Transactions on Image Processing}{Yu \MakeLowercase{\textit{et al.}}: Progressive Transfer Learning}
%,~Vol.~??, No.~?, ?????~????}%
\fi

\maketitle

\begin{abstract}
Model fine-tuning is a widely used transfer learning approach in person Re-identification (ReID) applications, which fine-tuning a pre-trained feature extraction model into the target scenario instead of training a model from scratch. It is challenging due to the significant variations inside the target scenario, e.g., different camera viewpoint, illumination changes, and occlusion. These variations result in a gap between the distribution of each mini-batch and the whole dataset's distribution when using mini-batch training. In this paper, we study model fine-tuning from the perspective of the aggregation and utilization of the global information of the dataset when using mini-batch training. Specifically, we introduce a novel network structure called Batch-related Convolutional Cell (BConv-Cell), which progressively collects the global information of the dataset into a latent state and uses it to rectify the extracted feature. Based on BConv-Cells, we further proposed the Progressive Transfer Learning (PTL) method to facilitate the model fine-tuning process by jointly optimizing the BConv-Cells and the pre-trained ReID model. Empirical experiments show that our proposal can improve the performance of the ReID model greatly on MSMT17, Market-1501, CUHK03 and DukeMTMC-reID datasets. Moreover, we extend our proposal to the general image classification task. The experiments in several image classification benchmark datasets demonstrate that our proposal can significantly improve the performance of baseline models. The code has been released at \url{https://github.com/ZJULearning/PTL}
\end{abstract}
\begin{IEEEkeywords}
Person Re-identification, Transfer Learning, Image Classification
\end{IEEEkeywords}

% \fi
\IEEEpeerreviewmaketitle
\section{Introduction}

Person re-identification (ReID) is to re-identify the same person in different images captured by different cameras or at different time. Due to its wide applications in surveillance and security, person ReID has attracted much interest from both academia and industry in recent years. 

With the development of deep learning methods and the newly emerged person ReID datasets, the performance of person ReID has been significantly boosted recently. However, several open problems remain. First, training a feature extraction model from scratch needs a large volume of annotation data. However, the annotated data is hard to acquire in person ReID tasks due to the poor quality of the image and pedestrians' privacy concerns. Hence, using existing datasets to help optimize the feature extractor has attracted great attention in the community. Second, the significant variations between different scenarios and within the same scenario make the person ReID task challenging. A noticeable performance degradation often occurs if we directly apply a pre-trained model on the target dataset without fine-tuning it into the target scenario.

Most of the recently proposed works \cite{deng2018image,ma2018disentangled} have focused on mitigating the impact of variations between different datasets. Most of these works focus on transferring the image style of the target domain and the source domain to the same by using Generative Adversarial Networks (GANs) based models. However, the imperfect style transferring models can bring in noises and potentially change the data distribution of the whole dataset. Meanwhile, the person ID in the generated images is not guaranteed to be the same as in the real images.

As for variations inside the dataset, which we focused on in this work, it is less mentioned in recently proposed works. The distribution difference between each mini-batch and the entire dataset caused by internal variations significantly influences the model fine-tuning process. This difference leads to a deviation of gradient estimation and thus affect the effect of model fine-tuning. The most straightforward approach to mitigate this problem is increasing the batch size. However, Keskar et al. \cite{DBLP:journals/corr/KeskarMNST16} and our experiments revealed that using a large-batch setting tends to converge to sharp minimizers, and further leads to poorer performance.

Moreover, most state-of-the-art deep learning methods in person ReID task have used an off-the-shelf network, like DenseNet \cite{huang2017densely} and  ResNet \cite{he2016deep}, as backbone network. However, Deep CNNs are difficult to initialize and optimize with limited training data. Therefore, model fine-tuning is widely used to mitigate shortages of annotated training data in person ReID tasks, making the study of how to mitigate the impact of the internal variation more critically. For instance, most of the off-the-shelf models used in ReID tasks are pre-trained on a relatively larger dataset like ImageNet \cite{ILSVRC15} and then fine-tuning into the target dataset. 

We study how to mitigate the impact of internal variations from the viewpoint of aggregation and utilization of the global information of the dataset. First, we propose a novel CNN building block, which we call the Batch-related Convolutional Cell (BConv-Cell). The BConv-Cell progressively aggregates the global information of the dataset into a latent state in a batch-wise manner. The latent state aggregated in previous batches will be used to mitigate the impact of the subsequent batches' internal variations. Based on the BConv-Cells, we further propose the Progressive Transfer Learning (PTL) method to fine-tune the pre-trained model by integrating it with the BConv-Cells. We conduct extensive experiments on MSMT17 \cite{wei2018cvpr}, Market-1501 \cite{zheng2015scalable}, CUHK03 \cite{li2014deepreid} and DukeMTMC-reID \cite{zheng2017unlabeled} datasets to show that our proposal can effectively promote the ReID performance.

Moreover, we propose a variation of the BConv-Cell (BConv-Cell-v2) in this work and extend the application scenario of our proposal from the person ReID task to general image classification tasks. Differ from the BConv-Cell, the BConv-Cell-v2 not only maintain the latent states of the BConv-Cells of previous mini-batches but also the output hidden states. We further use the BConv-Cell-v2 to replace the BConv-Cells we used in the PTL network and named it PTL-v2. The experimental results in several image classification benchmark datasets demonstrate that PTL-v2 outperforms baselines in all experiments, and outperforms the PTL in the dataset with fewer categories.

We summarize the contributions of this work as follows:
\begin{enumerate}
    \item We propose a novel network structure called the Batch-related Convolutional Cell (BConv-Cell) and its variation (BConv-Cell-v2). In mini-batch training, the BConv-Cells can progressively aggregate the global information of the dataset and then use it to optimize the model in the next batches. 
    \item Based on the BConv-Cells, we propose the Progressive Transfer Learning (PTL) method to fine-tune a pre-trained model into the target scenario by integrating the BConv-Cells. 
    \item The experimental results show that the model fine-tuned using our proposal can achieve state-of-the-art performance on four persuasive person ReID datasets.
    \item We extend the application scenario of our proposal from the person ReID task to general image classification tasks in this work. Experimental results in several image classification benchmark datasets demonstrate that our proposal can significantly improve the performance of the backbone models.
\end{enumerate}

\begin{figure*}[htp]
    \centering
    \begin{subfigure}
    \centering
    \includegraphics[width=.38\textwidth]{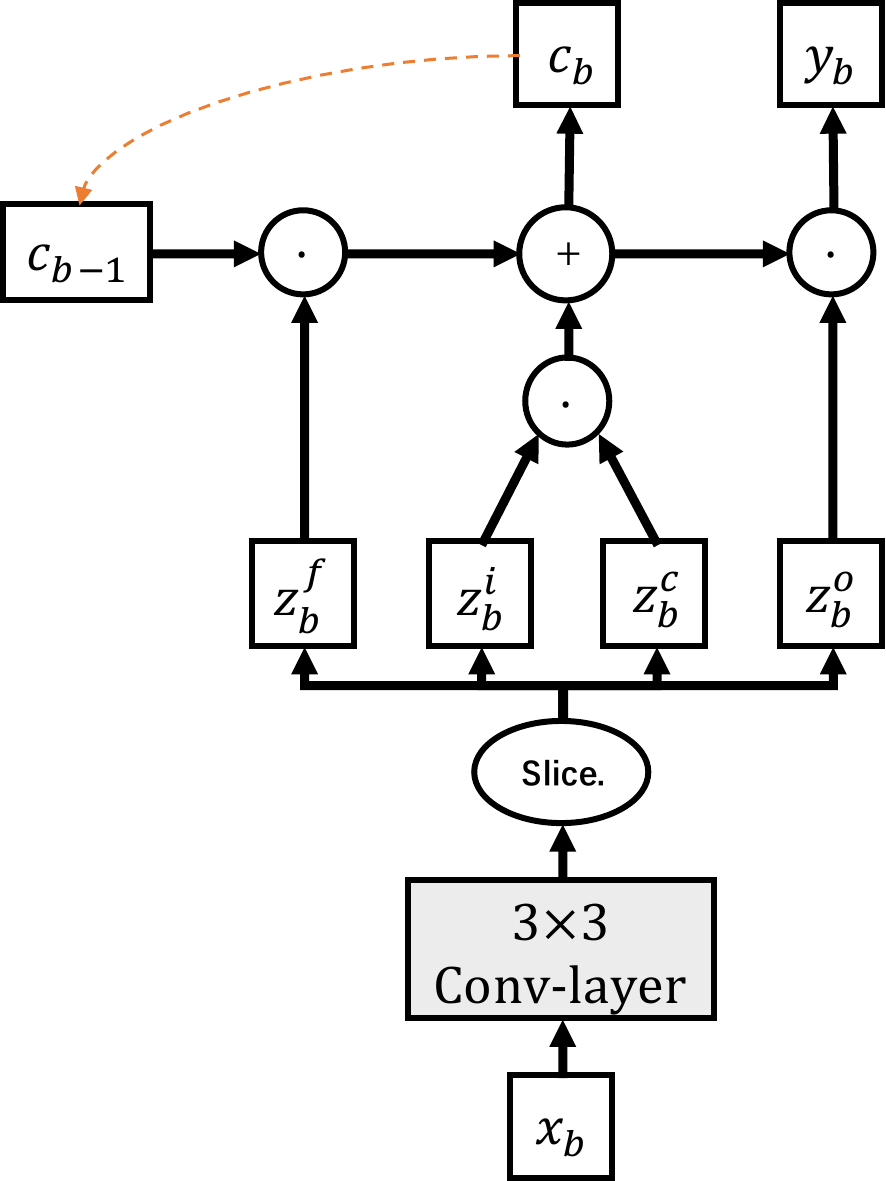}
    \end{subfigure}
    \begin{subfigure}
    \centering
    \includegraphics[width=.38\textwidth]{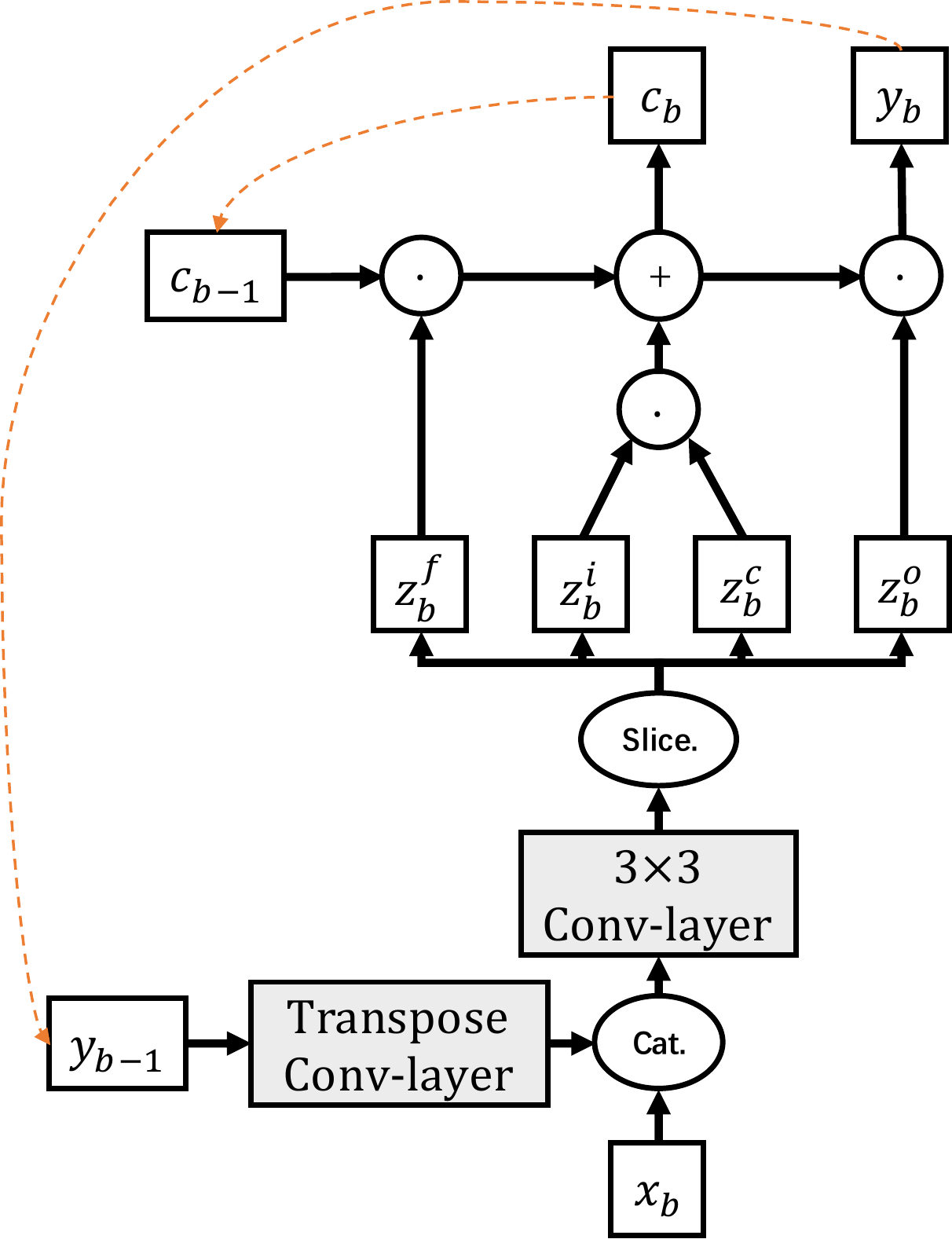}
    \end{subfigure}
    \caption[]{The network architecture comparison of two kind of BConv-Cells we used in PTL and PTL-v2. Left: BConv-Cell, which we used in PTL. Right: BConv-Cell-v2 used in PTL-v2, which we proposed in this work. The circle and ellipse in this figure denote operators, e.g., 'Cat.' denotes tensor concatenation operator. The hollow rectangle and square denote tensors, and the gray rectangles denote CNN layers.}
    \label{fig:bconv-compare}
\end{figure*}

\section{The Batch-related Convolutional Cell and its variation}
\label{subsection:bconv-cell}
The Batch-related Convolutional Cell (BConv-Cell) is based on a straightforward thought that making use of the global information of the dataset to mitigate the adverse influence caused by internal variation. In this section, we introduce the formula and architecture of the BConv-Cell and its variation BConv-Cell-v2.
\subsection{The Batch-related Convolutional Cell}
The BConv-Cell is inspired by the Conv-LSTMs \cite{xingjian2015convolutional}, as shown in the left of Figure \ref{fig:bconv-compare}. However, there are several fundamental differences between the BConv-Cells and the Conv-LSTMs. First, there is no time concept and explicit temporal connections between inputs in the BConv-Cells. Meanwhile, the BConv-Cells is not designed to handle sequential inputs but single segmented images. Second, the BConv-Cells have a different architecture from the Conv-LSTMs. The BConv-Cells only maintained a latent state that contained the aggerated global information, but the Conv-LSTMs reserved both the hidden state and the cell state. Moreover, the BConv-Cells is not designed to conduct prediction. Using the memory mechanism, the BConv-Cells can progressively collect global information and use it to facilitate the parameter optimization process during fine-tuning. Unlike other LSTM based methods like meta-learners, the output of the BConv-Cells can be directly used as the extracted feature. Meanwhile, the nature of the BConv-Cells is a stack of Conv-layers so that it can be used as a building block of a multi-layer feature extraction network. 
The key equations of the BConv-Cell have shown as follow:
\begin{equation}
\label{equ:bconvlstm}
\begin{aligned}
    & i_{b} = \sigma(W_{xi} \ast x_{b} + b_{i}) \\
    & f_{b} = \sigma(W_{xf} \ast x_{b} + b_{f})\\
    & o_{b} = \sigma(W_{xo} \ast x_{b} + b_{o})\\
    & C_{b} = f_{b} \circ C_{b-1} + i_{b} \circ tanh(W_{xc} \ast x_{b} + b_{c})\\
    & y_{b}= o_{b} \circ tanh(C_{b}),
\end{aligned}
\end{equation}
where $\ast$ denotes the convolution operator, $\circ$ denotes the Hadamard product, $\sigma$ denotes a sigmoid function, $x_b$ is the input of the BConv-Cell in $b$-th batch. $i_{b}$, $f_{b}$ and $o_{b}$ is the output of input gate $i$, forget gate $f$ and output gate $o$ respectively, $C_b$ is the latent state reserved after $b$-th batch, $W$ is the weight of the corresponding convolutional layer in the BConv-Cell and $y_{b}$ is the output of the BConv-Cell. All the input $x_b$, latent state $C_b$ and gate output $i_{b}, f_{b}, o_{b}$ are 3-dimensional tensors.

As shown in Eq.\,\ref{equ:bconvlstm}, the output $y_b$ is determined by the latent state $C_{b}$ and the input $x_b$. The latent state $C_{b}$ is determined by the input $x_b$ and $C_{b-1}$. From the fourth formula of Eq.\,\ref{equ:bconvlstm}, we can notice that the $C_{b}$ maintains part of the information of all the historical input batches. The iteration formula of latent state $C_{b}$ as:

\begin{equation}
\label{equ:cell-states}
C_{b} = g(x_{1},x_{2},...,x_{b}),
\end{equation}
where $g$ is the simplified notation of the composition of functions $\{g_{i} | 1 \leq i \leq b \}$.

\subsection{The Batch-related Convolutional Cell-v2}

In this work, we also propose a variation of the BConv-Cell and name it BConv-Cell-v2. The comparison of the BConv-Cell and BConv-Cell-v2 have shown in Figure \ref{fig:bconv-compare}. As we can notice, BConv-Cell-v2 maintains the latent states of previous mini-batches and the output hidden states at the same time. In each mini-batch, the output hidden states are aggregated with the hidden states of previous mini-batches. Meanwhile, the features of historical inputs are partially recovered by using a transpose convolution layer before concatenated with the input features of the current mini-batch. We argue that the concatenated features of historical inputs can mitigate the impact of the data generation bias of mini-batch, and thus improve the stability of the model. More explanation can be found in Section \ref{sec:CbE}.

\section{Progressive Transfer Learning Network}
\label{section:ptl-network}
\subsection{Progressive Transfer Learning}
\begin{figure*}[ht]
    \centering
    \includegraphics[width=\textwidth]{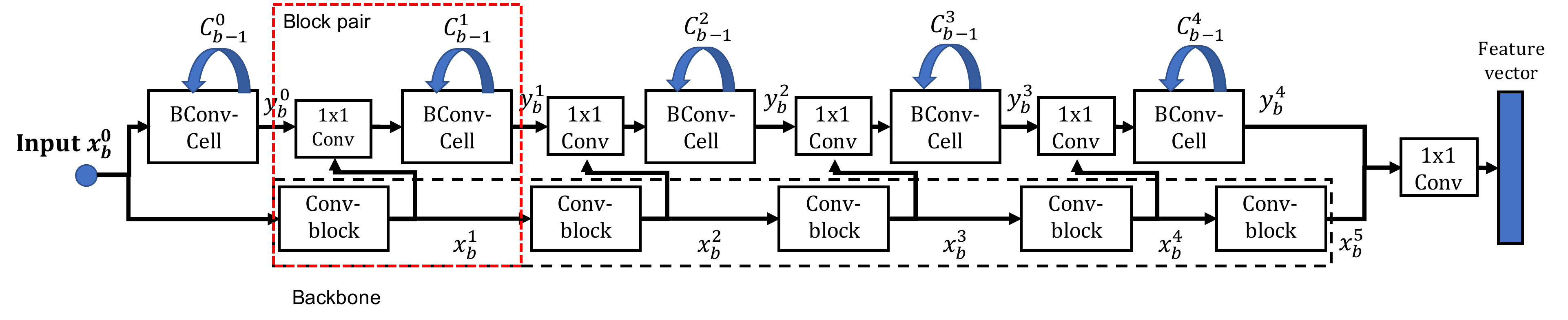}
    \caption[]{Sketch of the PTL network. The black dotted box indicates the backbone network. $x_{b}^{i}$ and $y_{b}^{i}$ are the outputs of the $i$-th Conv-block of the backbone and the related BConv-Cell respectively, $x_{b}^{0}$ denotes the input image, $b$ indicates the $b$-th input batch, $C_{b-1}^{i}$ is the latent state of the $i$-th BConv-Cell after the last batch. The red dotted box denotes the block pair of the Conv-block and the BConv-Cell. In each block pair, $x_{b}^{i}$ and $y_{b}^{i-1}$ are concatenated and feeding into a 1x1 Conv layer before feed into the BConv-Cell. The output of the last BConv-Cell and that of backbone network are concatenated before feeding into the 1x1 Conv-layer. The latent state of the BConv-Cell is stored after every batch and feedback to the same BConv-Cell when next batch coming.}
    \label{fig:1}
\end{figure*}

Given an off-the-shelf CNN as the backbone network, we pair up the BConv-Cells with its Conv-blocks to form a new network, and we name it as the progressive transfer learning (PTL) network. A sketch of the PTL network has shown in Figure \ref{fig:1}. The red dotted box denotes a building block of the PTL network, which formed by a BConv-Cell, a 1x1 Conv-layer and the Conv-block of the backbone network. Formally, we define this building block as:
\begin{equation}
\label{equ:inference}
\begin{aligned}
    x_{b}^{i} &= F_{conv}(x_{b}^{i-1}) \\
    y_{b}^{i} &= F_{bconv}(F_{1\times1}(x_{b}^{i}, y_{b}^{i-1}), C_{b-1}^{i}) \\
    C_{b}^{i} &= g(x_{1}^{i}, x_{2}^{i}, ..., x_{b}^{i}),
\end{aligned}
\end{equation}
where $x_{b}^{0}$ indicate the input image of $b$-th batch, $x_{b}^{i}$ is the output of the $i$-th $(i>=1)$ Conv-block in $b$-th batch, $y_{b}^{i}$ is the output of the $i$-th BConv-Cell. Eq.\,\ref{equ:inference} only contains the second and the third equation when $i=0$. The function $F_{conv}$ and $F_{bconv}$ represent the mapping function learned by Conv-block and BConv-Cell respectively. $F_{1\times1}$ is the 1x1 Conv-layer as shown in Figure \ref{fig:1}. $C_{b}^{i}$ is the latent state of the $i$-th BConv-Cell after the $b$-th batch. The structure of the Conv-block is flexible, which can be replaced by Conv-block of many Deep CNNs like DenseNet or ResNet. 

As shown in Eq.\,\ref{equ:inference}, the BConv-Cell learn the mapping function from input to feature space while collecting global information and updating the latent state. We can notice from Eq.\,\ref{equ:inference} that the discriminative knowledge of the past batches is progressively aggregated into the latent state.

\subsection{Progressive Transfer Learning Network Architecture}
\label{section: network-arch}

We have tested the PTL method with several different structures of backbone networks, including DenseNets and ResNets. We use the DenseNet-161 as the backbone network to describe the construction of the PTL network. 

The DenseNet-161 consists of five Conv-blocks. We use four BConv-Cells to pair up with the top four Conv-blocks, as shown in Figure \ref{fig:1}. At the top of the network, we use a BConv-Cell to capture the low-level feature of the input image, as shown in the left of Figure \ref{fig:1}. At the bottom of the network, the output of the last BConv-Cell is concatenated with the output of the last Conv-block and then feed into a 1x1 Conv-layer to get the feature vector. During training, the feature vector is then fed into a classifier which contains three Fully connection layers. For simplicity, the classifier is not shown in Figure \ref{fig:1}. During evaluating, we directly use the feature vector conduct image retrieve.

As we can see in Figure \ref{fig:1}, feature maps transmit along two dimensions in the PTL network. The first is batch iteration, BConv-Cells evolve the latent states with each input batch and transmit it to the next batch. The second is the depth of the network, in which feature maps transmit from the first layer to the last layer.

During testing, we set all the latent states as zeros.  We set all the latent states to zeros at the beginning of each epoch during training to simulate the test condition. This setting ensures that historical knowledge is progressively collected and aggregated only once in each epoch.

As mentioned above, the backbone in Figure \ref{fig:1} can be replaced by most of the commonly used feature extraction networks. In this work, we use ResNet-50, DenseNet-161 and MGN \cite{wang2018learning} as backbone network.

\subsection{Progressive Transfer Learning Network-v2}
In this work, we also propose the PTL-v2 network, a variation of PTL network in which the BConv-Cells are replaced with BConv-Cell-v2 blocks, while all other components remain unchanged. The difference between BConv-Cell and  BConv-Cell-v2 has shown in Figure \ref{fig:bconv-compare}. 

In each BConv-Cell-v2 block of PTL-v2, we first recover the features of historical inputs of previous mini-batches before concatenating it with the current mini-batch input features. We use the aggregated information of previous mini-batches to disturb the newly input features, thus mitigate the impact of the data generation bias. The experimental results show that the PTL-v2 can outperform all baselines in several image classification benchmarks while outperforming the PTL network in image classification benchmarks with fewer categories. \cite{ijcai2020-345}

\subsection{Parameter Optimization Procedure}
\label{subsection:PTL based Deep Optimization}
Our proposal facilitates parameter optimization using the BConv-Cells to cooperate with the backbone network, which has no limit on the model optimization method. Hence, the combined model still can be optimized by using commonly used optimizers like SGD and SGD-M.

We argue that the PTL method can make up for two shortcomings of the SGD-M optimizer. First, in SGD-M, the historical gradient is aggregated in a linear sum roughly by using humanly pre-defined weights, which make it inflexible and not optimized. Second, the loss after each batch only determined by the current input batch, which has a strong bias and leading to performance oscillation during training. 

Using the PTL method, the historical gradient aggregation is replaced by calculating the gradient of a composition function recursively with learnable weights. More than that, the sample bias of the current batch can be mitigated by using the historical knowledge carried by the learned latent states $C_{b}$.

\subsection{Another Causality based Explanation}
\label{sec:CbE}
We also offer a causality-based explanation to help understand why our approach works. Many deep-learning based methods typically follow a consistent assumption that the training and testing datasets are generated from the same data distribution (i.e., i.i.d hypothesis). Under i.i.d hypothesis, a trained model can be directly applied to the testing dataset, and its performance should be equivalent to the training dataset. Although this approach is empirically proven to be highly successful in many public datasets, it is considered flawed in real-world applications. 

In real-world applications, we rarely know the true underlying model, and we cannot guarantee the unknown test data will have the same distribution as the training data. Moreover, in real-world applications, significant data generation bias can result in substantial variations between different training and testing datasets. Consequently, models trained in such a training dataset will fall into the statistical correlation of features and result in unstable performance to yielding lower training loss. To mitigate this problem, our proposal progressively aggregates the global information of the training dataset and then applying it to mitigate the impacts of data generation bias during training. As shown in experiments, our proposal can significantly improve the performance of backbone models in both person ReID and general image classification tasks.

\begin{figure*}[!tp]
    \centering
    \begin{subfigure}
    \centering
    \includegraphics[width=.45\textwidth]{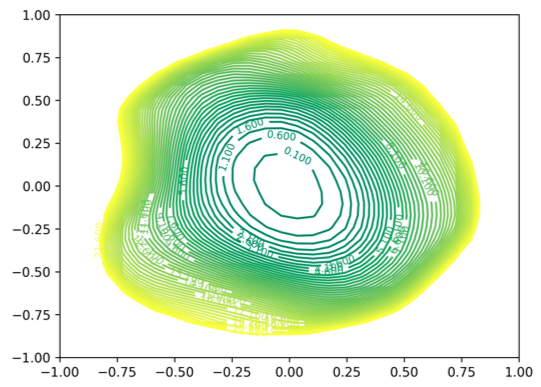}
    \end{subfigure}
    \begin{subfigure}
    \centering
    \includegraphics[width=.45\textwidth]{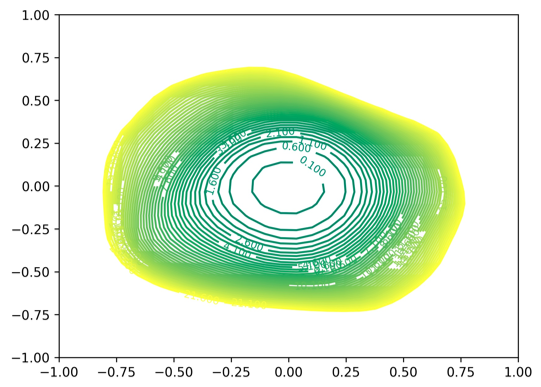}
    \end{subfigure}
    \caption[]{The loss surfaces of DenseNet-100 with/without PTL-v2 on CIFAR-100 dataset generated by using visualization tools proposed by \protect \textit{Li et al} \cite{visualloss}. Left: DenseNet-100+PTL-v2, Right: DenseNet-100. The large the loss surface, the better the model's generalization ability. }
    \label{fig:4}
\end{figure*}

To verify our idea, we further visualized the loss surfaces of DenseNet-100 with/without PTL-v2 on the CIFAR-100 dataset using the visualization tools proposed by \textit{Li et al.} \cite{visualloss}. As shown in \textbf{Figure \ref{fig:4}}, we can notice that the loss surface of DenseNet-100+PTL-v2 is wider than DenseNet-100. As \textit{Li et al.} \cite{visualloss} illustrated and proved in their work, the large the loss surface, the better the model's generalization ability. Therefore, we can draw a plain conclusion that PTL helps improve the model's performance in the fine-tuning process using the global information of the dataset.

\section{Student-Teacher Distillation Method}
\label{section:STD}
\begin{figure}[!tp]
    \centering
    \includegraphics[width=0.5\textwidth]{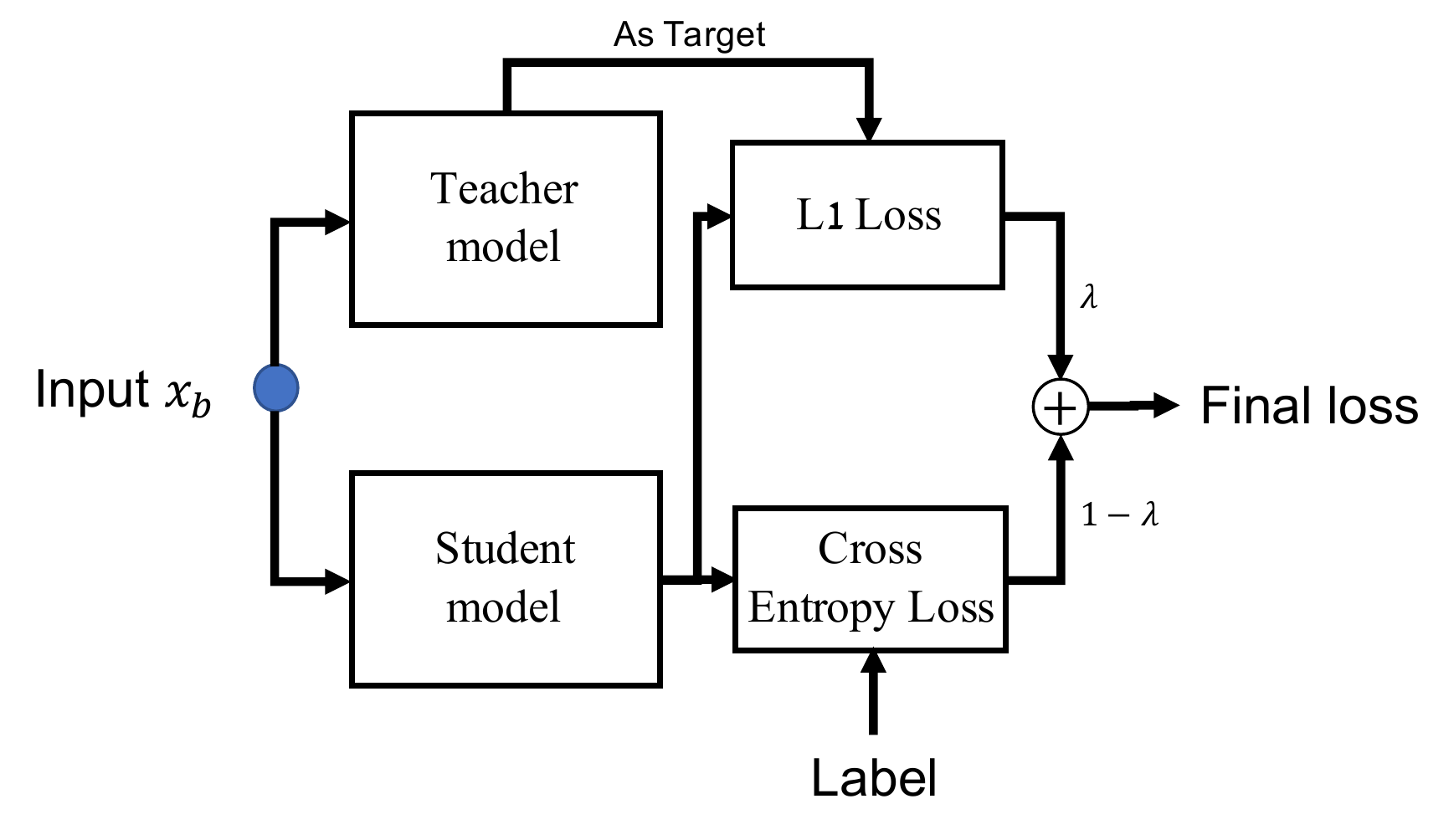}
    \caption[]{The implementation of STD method. The teacher model is set to evaluation mode during the whole process.}
    \label{fig:3}
\end{figure}
Compared with the backbone network, the parameter number of the PTL network grew up inevitably. To fairly compare with baselines, we introduce an improved model distillation method called Student-Teacher Distillation (STD) method to fine-tune a backbone model (like ResNet-50) by using the fine-tuned PTL model. The STD method is not essential for the PTL method in practical applications. 

As the prerequisite, we assume we have obtained a fine-tuned model by using our proposed PTL method. We then introduce a new objective function for model distillation. The objective function consists of two parts. First is a cross-entropy loss between the output predictions of the student model and the ground truth. The second is an L1 loss between the student model's output feature vector and that of the teacher model. The new objective function is given by:
\begin{equation}
\label{equ:std}
    L_{distill} = (1-\lambda) L_{CE} + \lambda L_{l1},
\end{equation}
where $\lambda$ is a hyper-parameter to adjust the ratio of the cross-entropy loss and the L1 loss. This new object function combines both supervision information and merit of the PTL network to extract discriminative features. 

The implementation of the STD method has shown in Figure \ref{fig:3}. We set the teacher model to evaluation mode during the whole process. The input image feeds into teacher and student model at the same time. After this, the parameter of the student network will be updated according to the proposed objective function in Eq.\,\ref{equ:std}. After training, the teacher model can be abandoned.

\section{Towards General Image Classification}
We further applied our proposal to general image classification tasks. The key idea of PTLs of progressively aggregating the global information of the dataset is adaptable in general image classification tasks. 

In a typical person ReID task, we first  obtained the embeddings of all gallery and query images using the feature extraction model. After that, metric learning methods are used to learn a distance metric. Then, the nearest neighbor of the query image is retrieved from the gallery set, and its label is output as prediction label for the query image. In a general image classification task, we followed the standard image classification procedure. The output of the feature extraction model (PTL or PTL-v2) is fed into a classifier that consists of two linear layers. The output of the classifier is then fed into a Softmax function to generate a predicted class label.

\section{Experiments}
\label{section:experiments}
We first carried out model fine-tuning experiments with our proposal on four convincing ReID datasets and compared it with both the state-of-the-art ReID methods and several transfer-learning methods. We then conduct model transferring experiments among multiple datasets to evaluate the PTL method's performance when handling multiple-step transferring. 

\subsection{Dataset}

We selected four persuasive ReID datasets to evaluate our proposal, including Market-1501, DukeMTMC-reID, MSMT17, and CUHK03. We further applied our proposal to general image classification tasks, and tested it in several image classification datasets: CIFAR-10, CIFAR-100.

\paragraph{Market-1501.} The Market-1501 dataset contains 32,668 annotated bounding boxes of 1,501 identities. 

\paragraph{DukeMTMC-reID.} The DukeMTMC-reID dataset contains 1,404 identities. 702 IDs are selected as the training set and the remaining 702 IDs as the testing set.

\paragraph{MSMT17.} The raw video on the MSMT17 dataset is recorded in 4 days with different weather conditions in a month using 12 outdoor cameras and three indoor cameras. The MSMT17 dataset contains 126,441 bounding boxes of 4,101 identities. We followed the same dataset split by Wei et al. \cite{wei2018cvpr}, and we also used the evaluation code provided by them (\url{https://github.com/JoinWei-PKU/MSMT17_Evaluation}). 
\paragraph{CUHK03.} The CUHK03 dataset consists of 14,097 images of
1,467 persons from 6 cameras. Two types of annotations
are provided in this dataset: manually labeled pedestrian
bounding boxes and DPM-detected bounding boxes. We followed the same dataset split as used in the \cite{wang2018learning}.  For all experiments on Market-1501, DukeMTMC-reID and CUHK03, we used the evaluation code provided in Open-ReID (\url{https://github.com/Cysu/open-reid}).

\paragraph{Market-Duke.} We use the training sets of the two datasets Market-1501 and DukeMTMC-reID to form a new dataset called the Market-Duke dataset. We further use this dataset to train the models to compare the difference between the one-step model fine-tuning and multi-step model fine-tuning. All validation, query and gallery set of these two datasets are abandoned.

\paragraph{CIFAR-10.} The CIFAR-10 dataset consists of 60000 32x32 color images in 10 classes, with 6000 images per category. There are 50000 training images and 10000 test images.

\paragraph{CIFAR-100.} This dataset is just like the CIFAR-10, except it has 100 classes containing 600 images each. There are 500 training images and 100 testing images per category. The 100 classes in the CIFAR-100 are grouped into 20 superclasses.

\subsection{Experiment Setting}
\begin{table}[t]
\centering
\caption[]{Results on the MSMT17 dataset.}
\resizebox{\columnwidth}{!}{
\begin{tabular}{lccc}
\toprule
\textbf{Method}              & \textbf{\#Param.}$^1$ & \textbf{mAP}   & \textbf{CMC-1} \\ \midrule
GoogLeNet \cite{wei2018cvpr}    & -          & 23.00          & 47.60                          \\

PDC \cite{wei2018cvpr}        & -          & 29.70          & 58.00                    \\
GLAD \cite{wei2018cvpr}      & -          & 34.00          & 61.40                   \\
 \midrule
ResNet-50    & 28m   & 28.63          & 59.77                       \\
ResNet-50+PTL   & 35m   & 32.58          & 62.76                 \\
DenseNet-161 & 32m   & 38.60          & 70.80                               \\
DenseNet-161+PTL & 42m   & \textbf{42.25} & 72.65  \\
DenseNet-161+PTL+STD & 32m   & 41.38 & \textbf{73.12} \\ \bottomrule
\end{tabular}}
\begin{tablenotes}
\small
\item 1. \#Param. indicates parameter number, while $m$ indicates million.
\end{tablenotes}
\label{tab:randomstart-msmt}
\end{table}
We select the DenseNet-161 model and ResNet-50 model both pre-trained on the ImageNet dataset as the backbone model. As for the state-of-the-art model in ReID tasks, we select the MGN \cite{wang2018learning} model, which also uses a ResNet-50 as the backbone network. We modified the backbone network by using our proposed PTL method, and name these models as DenseNet-161+PTL, ResNet-50+PTL and MGN+PTL, respectively. We then use the STD method to train the DenseNet-161 model (DenseNet-161+PTL+STD) using the DenseNet-161+PTL as the teacher model.

In person ReID experiments, images have been reshaped into 256x128 (height x width) before feeding into the network except for the experiments of MGN and MGN+PTL, which use image size 384x128. We take out the output of the 1x1 Conv-layer as the discriminative feature. The initial learning rate is set to 0.01 and decay the learning rate ten times every ten epochs. Models are fine-tuned for 50 epochs. Unless otherwise stated, in all of our experiments, we use SGD-M as the optimizer. The hyper-parameter $\lambda$ is set to 0.8 by practicing in the following experiments. For experiments involved MGN and MGN+PTL, we followed the experiment setting in \cite{wang2018learning}. Meanwhile, we use the single-query setting in all experiments.

To the best of our knowledge, the official implementation of the MGN has not been released. Hence, we use the reproduction version published in \url{https://github.com/GNAYUOHZ/ReID-MGN}. The results obtained by this reproduction code are noted as MGN (reproduced) in all the tables. Although the MGN model uses the ResNet-50 as the backbone network, the parameter number of the MGN model (66m) is much more than the ResNet-50 model (28m). Due to the GPU usage limitation, we have not conducted experiments about the MGN+PTL+STD.

We use the cross-entropy loss in all of the fine-tuning processes in our experiments, except for the MGN+PTL. The MGN uses a combined loss function insists of cross-entropy loss and triplet loss, we use the same loss function in all the experiments of the MGN+PTL.

\subsection{MSMT17}

We first evaluate the PTL method on the MSMT17 dataset. The detailed evaluation statistics has shown in Table \ref{tab:randomstart-msmt}. We can see a significant performance promotion by using the PTL method. The performance of the DenseNet-161+PTL+STD outperforms not only the backbone model but also all of the baseline methods. We also can notice that the DenseNet-161+PTL+STD model can achieve a higher CMC-1 score than the DenseNet-161+PTL model. We attribute the success to the combined loss function of the STD method. By combining the cross-entropy Loss with the L1 loss, the student model can learn the discriminative knowledge from the teacher model while imposing restrictions on the learned knowledge.

\subsection{Market-1501}
\begin{table}
\centering
\caption[]{Results on the Market-1501 dataset.}
\begin{tabular}{lcc}
\toprule
\textbf{Method}        & \textbf{mAP}   & \textbf{CMC-1}   \\ \midrule
DML \cite{zhang2018deep} & 70.51          & 89.34                    \\
HA-CNN \cite{li2018harmonious}   & 75.70          & 91.20               \\
PCB+RPP \cite{sun2018beyond}      & 81.60       & 93.80             \\
MGN \cite{wang2018learning} & 86.90 &95.70 \\
\midrule
DenseNet-161*$^1$    & 69.90          & 88.30                \\
DenseNet-161    & 76.40          & 91.70            \\
DenseNet-161+PTL  & 77.50          & 92.50                \\
DenseNet-161+PTL+STD    & 77.50          & 92.20      \\
MGN (reproduced)$^2$   & 85.80       & 94.60      \\ 
MGN+PTL   & \textbf{87.34}          & \textbf{94.83} \\\bottomrule
\end{tabular}
\begin{tablenotes}
\item \small{1. DenseNet-161* used a batch size of 90, while DenseNet-161 used batch size 32.}
\item \small{2. MGN (reproduced) is our reproduction of the MGN \cite{wang2018learning}.}
\end{tablenotes}
\label{tab:market}
\end{table}
We use DenseNet-161 and MGN as backbone model to evaluate the performance of the PTL method on Market-1501 dataset. We select several state-of-the-art person ReID methods as baselines. Among these methods, the DML \cite{zhang2018deep} is also pre-trained on ImageNet and transferred to Market-1501. 

The results has summarized in Table \ref{tab:market}. We can notice that a simple DenseNet-161 model can outperform the state-of-the-art transfer learning based person ReID methods on the Market-1501 dataset using the PTL method and the STD method. Meanwhile, the MGN+PTL outperforms all the state-of-the-art methods.

Moreover, we can notice that using a large batch size (DenseNet-161*) is not an effective way to narrow the gap between the distribution of each mini-batch and the distribution of the dataset. In contrast, large batch size can result in poor performance.

\subsection{CUHK03}
We then conduct model fine-tuning experiments on the CUHK03 dataset. We compare the performance of MGN+PTL with several state-of-the-art methods. The results have summarized in Table \ref{tab:cuhk03}. We can notice that by using the PTL method, the ReID performance of the MGN model has promoted tremendously, and outperforms all the state-of-the-art methods.

\begin{table}
\centering
\caption[]{Results on the CUHK03 dataset.}
\resizebox{\columnwidth}{!}{
\begin{tabular}{lcccc}
\toprule
Methods             & \multicolumn{2}{c}{Detected} & \multicolumn{2}{c}{Labelled}    \\
                                                & mAP               & CMC-1             & mAP            & CMC-1          \\ \midrule
HA-CNN \cite{li2018harmonious} & 38.60             & 41.70             & 41.00          & 44.40          \\
PCB \cite{sun2018beyond}       & 54.20             & 61.30             & -              & -              \\
PCB+RPP \cite{sun2018beyond}  & 57.50             & 63.70             & -              & -              \\
MGN \cite{wang2018learning} & 66.00             & 66.80             & 67.40          & 68.00          \\
\midrule
MGN (reproduced)                                        & 69.41             & 71.64             & 72.96          & 74.07          \\
MGN+PTL                                         & \textbf{74.22}    & \textbf{76.14}    & \textbf{77.31} & \textbf{79.79} \\ \bottomrule
\end{tabular}}
\label{tab:cuhk03}
\end{table}

\subsection{DukeMTMC-reID}
\begin{table}
\centering
\caption{Results on the DukeMTMC-reID dataset.}
\begin{tabular}{lcc}
\toprule
\textbf{Method}        & \textbf{mAP}   & \textbf{CMC-1} \\ \midrule
HA-CNN \cite{li2018harmonious} & 63.80&80.50\\
PCB \cite{sun2018beyond} & 69.20&83.30\\
MGN \cite{wang2018learning}   & 78.40       & 88.70           \\ 
 \midrule
MGN (reproduced)   & 77.07       & 87.70         \\ 
MGN+PTL                                                                                                           & \textbf{79.16}          & \textbf{89.36}      

\\\bottomrule
\end{tabular}
\label{tab:duke}
\end{table}
We then conduct experiments on the DukeMTMC-reID dataset. As for baselines, we compare the performance with several state-of-the-art methods, including HA-CNN, PCB and MGN. The results have shown in Table \ref{tab:duke}, we can notice that by using our method, the MGN+PTL model can outperform all state-of-the-art methods.

\subsection{Transfer among Multiple Datasets}
In real-world applications, the ReID model needs to transfer among a sort of datasets to take advantage of all available data. Therefore, we conduct multiple datasets transferring experiments to evaluate the performance of our proposal when dealing with model fine-tuning among multiple datasets.  Similar to the experiment on the MSMT17 dataset, we also use the STD method to train a DenseNet-161 model to compare with the baselines fairly.

The detailed results have shown in Table \ref{tab:mutiple-trans}, from which we can see that the PTL method achieves better performance compared with baseline models. We also notice that the performance of two-step transferring achieves better performance compare with one step transferring. For instance, fine-tuning follow the order 'Duke to Market to MSMT17' outperforms 'Market-Duke to MSMT17'. We argue that the substantial style variation in the Market-Duke dataset is caused by the substantial style variation, which is richer than either Market-1501 or the DukeMTMC-reID dataset.

More than that, we can see that the order of fine-tuning can influence the performance of the final model. The model fine-tuned by 'Duke to Market to MSMT17' can achieve highest score in both mAP and CMC-1.
\begin{table}
\begin{center}
\caption[]{Results of transfer among multiple datasets.}
\resizebox{\columnwidth}{!}{
\begin{tabular}{llcc}
\toprule
\textbf{Method}  & \textbf{Fine-tuning list}*                                                        & \textbf{mAP}                    & \textbf{CMC-1}                 \\

\midrule
DenseNet-161     & Market-Duke to  MSMT17                & 41.12                           & 72.49                                                       \\
DenseNet-161+PTL+STD & Market-Duke to  MSMT17                & 42.53 & 74.11\\ \midrule
DenseNet-161     & Market to  Duke to  MSMT17 & 41.22                           & 72.78                                                               \\
DenseNet-161+PTL+STD & Market to  Duke to  MSMT17 & 42.34                  & 73.60                               \\\midrule
DenseNet-161     & Duke to Market  to  MSMT17 & 41.80                           & 73.00                                                           \\
DenseNet-161+PTL+STD & Duke to Market  to  MSMT17 & \textbf{42.73}                  & \textbf{74.31}                                \\ \midrule
DenseNet-161+PTL & Duke to Market                                                        & 76.00 & 91.30          \\
DenseNet-161+PTL+STD & Duke to Market                                                        & 75.50 & 91.10    \\\midrule
DenseNet-161+PTL & Duke to MSMT17 to Market                                                     & \textbf{77.90}  & 91.60         \\
DenseNet-161+PTL+STD & Duke to MSMT17 to Market                                                      & 77.40 & 91.60           \\\bottomrule
\end{tabular}}
\end{center}
\begin{tablenotes}
\small
\item *Fine-tuning list indicates the fine-tuning order. E.g., 'Duke to Market' means the model is fine-tuned on DukeMTMC-reID before fine-tune it on Market-1501, while 'Market' and 'Duke' denote Market-1501 dataset and DukeMTMC-reID dataset respectively.
\end{tablenotes}
\label{tab:mutiple-trans}
\end{table}

\subsection{Evaluate STD Method on MSMT17}
In this subsection, we evaluate the STD method on the MSMT17 dataset. We conduct a series of comparative experiments by adjusting the ratio of the cross-entropy loss and L1 loss.

We use the DenseNet-161+PTL model transferring from Market-1501 to MSMT17 in Table \ref{tab:mutiple-trans} as the teacher model. The student model is a DenseNet-161 model which has been transferred from ImageNet to Market-1501 using a SGD-M optimizer.
\begin{table}
\centering
\caption[]{Results of the STD method on the MSMT17 dataset.}
\resizebox{\columnwidth}{!}{
\begin{tabular}{lccccc}
\toprule
\textbf{Method}      & \textbf{\#Param.} & $\mathbf{\lambda}$* & \textbf{mAP}   & \textbf{CMC-1}   \\ \midrule
DenseNet-161+PTL     & 42m             & -                 & 42.45          & 72.48                       \\
DenseNet-161+PTL+STD & 32m             & 0.00              & 38.60          & 70.80                        \\
DenseNet-161+PTL+STD & 32m             & 0.30              & 41.26          & 72.52                \\
DenseNet-161+PTL+STD & 32m             & 0.50              & 42.27          & \textbf{73.49}              \\
%DenseNet-161+PTL+STD & 32012089             & 0.60              & 42.39          & 73.18          & 84.18                   \\
DenseNet-161+PTL+STD & 32m             & 0.80              & \textbf{42.51} & 73.37                        \\
%DenseNet-161+PTL+STD & 32012089             & 0.90              & 42.36          & 73.24          & 84.21              \\
DenseNet-161+PTL+STD & 32m             & 1.00              & 41.66          & 72.32                           \\ \bottomrule
\end{tabular}}
\begin{tablenotes}
\small
\item *$\lambda$ is the hyper-parameter in Eq.\,\ref{equ:std}, which denotes the proportion of L1 loss in the combined loss function. $\lambda = 0$ means use a SGD-M optimizer to fine-tune a DenseNet-161 model on MSMT17 without using the STD method.
\end{tablenotes}
\label{tab:std}
\end{table}

The detailed results are shown in Table \ref{tab:std}. From this table, we can see that by using the STD method, the performance of the DenseNet-161 model is promoted significantly. Meanwhile, we can see that the score of the DenseNet-161+PTL+STD grows up when $\lambda$ grows up. However, when $\lambda$ bigger than 0.8, the score no longer increases anymore. We argue that it is because the cross-entropy loss in the combined loss function is essential.

\subsection{General Image Classification Benchmarks}
In this work, we extend the application scenario of our proposal from the person Re-ID task to the general image classification tasks. Thus, we conduct experiments on three image classification benchmarks, including CIFAR-10, CIFAR-100, and CIFAR-100+ datasets. We compare the performance of the proposed PTL and PTL-v2 with the baseline model to evaluate our proposal's performance in general image classification tasks. The "+" indicates standard data augmentation (translation and/or mirroring). The detailed results are shown in Table \ref{tab:cifar}. From this table, we can notice that the PTL and PTL-v2 models' performance outperform the baseline model. Moreover, we can notice that the PTL-v2 model can outperform the PTL model in the CIFAR-10 dataset, which has fewer categories than the CIFAR-100 dataset.

\section{Related Works}
\label{section:relatedworks}
\subsection{Transfer Learning Methods}
Many transfer learning methods have been proposed recently. Zhong et al. \cite{DBLP:journals/corr/abs-1711-10295} proposed a domain adaption approach that transfers images from one camera to the style of another camera. Fan et al. \cite{fan2018unsupervised} proposed an unsupervised fine-tuning approach that used an IDE model trained on DukeMTMC-reID as the start point and fine-tuned it on target dataset. Unlike these approaches, our method is based on model fine-tuning, which is more flexible and easy to conduct. 

As for optimization methods used in transfer learning, methods in which a meta-learner is applied to learn how to update the parameters of the backbone model has attracted lots of attention recently \cite{DBLP:journals/corr/HaDL16,finn2017model}. In these approaches, parameters are updated using a learned update algorithm. For instance, Finn et al. \cite{finn2017model} proposed a meta-learning method MAML by using an LSTM network to update parameters. Our proposal is distinct from these approaches in several aspects. First, these meta-learning works aim to find a better parameter optimization route that can efficiently optimize the model parameter. Differently, the PTL network is designed to mitigate the distribution difference between mini-batch and the whole dataset. Meanwhile, the BConv-Cells can be directly participating in the feature extraction. 

\begin{table}
\centering
\caption{Error rates on CIFAR-10, CIFAR-100 and CIFAR-100+ datasets.}
\begin{tabular}{lccc}
\toprule
\textbf{Method}        & \textbf{CIFAR-10} & \textbf{CIFAR-100} & \textbf{CIFAR-100+}  \\ \midrule
DenseNet-100 & 5.92\% & 24.15\% & 22.27\% \\
DenseNet-100+PTL   & 5.62\%  & \textbf{23.17\%} & \textbf{21.52\%}    \\ 
DenseNet-100+PTL-v2  & \textbf{5.38\%}  & 23.2\% & 21.83\%     
\\\bottomrule
\end{tabular}
\label{tab:cifar}
\end{table}

\subsection{Person Re-identification Networks}
With the prosperity of deep learning, using deep learning networks as feature extractor has become common in person ReID tasks. Many deep learning based person ReID methods \cite{varior2016gated,zhang2017alignedreid,li2014deepreid} have been proposed. As for the transfer learning based deep person ReID method, Geng et al. \cite{geng2016deep} proposed a deep transfer learning model to address the data sparsity problem.

\subsection{Knowledge Distillation Methods}
Our proposed STD method is a special case of knowledge distillation \cite{hinton2015distilling}. More generally, it can be seen as a special case of learning with privileged information. Using distillation for model compression is mentioned by Hinton et al. \cite{hinton2015distilling}. Wu et al. \cite{DBLP:journals/corr/Wu16b} used the distillation method to improve the accuracy of binary neural networks on ImageNet.

\section{Conclusion}
\label{section:conclusion}
In this paper, we propose a Batch-related Convolutional Cell (BConv-Cell) and its variation BConv-Cell-v2 to mitigate the impact of the bias of each mini-batch caused by internal variations. The BConv-Cells can progressively collect the global information of the dataset during training while participating in the feature extraction. This global information will be used to mitigate the bias of each mini-batch in the next iterations. Based on the BConv-Cells, we propose the Progressive Transfer Learning (PTL) method to fine-tune the pre-trained model into the target dataset. Extensive experiments show that our method can significantly improve the backbone network's performance and achieve state-of-the-art performance on four datasets, including Market-1501, MSMT17, CUHK03, and DukeMTMC-reID datasets. Moreover, we extend our proposal's application scenario from the person Re-ID task to the general image classification tasks. We also evaluate the performance of our proposal in several image classification benchmarks. The experimental results show that our proposal can improve the performance of backbone models in image classification tasks.

\bibliographystyle{IEEEtran}
\bibliography{IEEEabrv,reference.bib}
\begin{IEEEbiography}
[{\includegraphics[width=1in,height=1.25in]{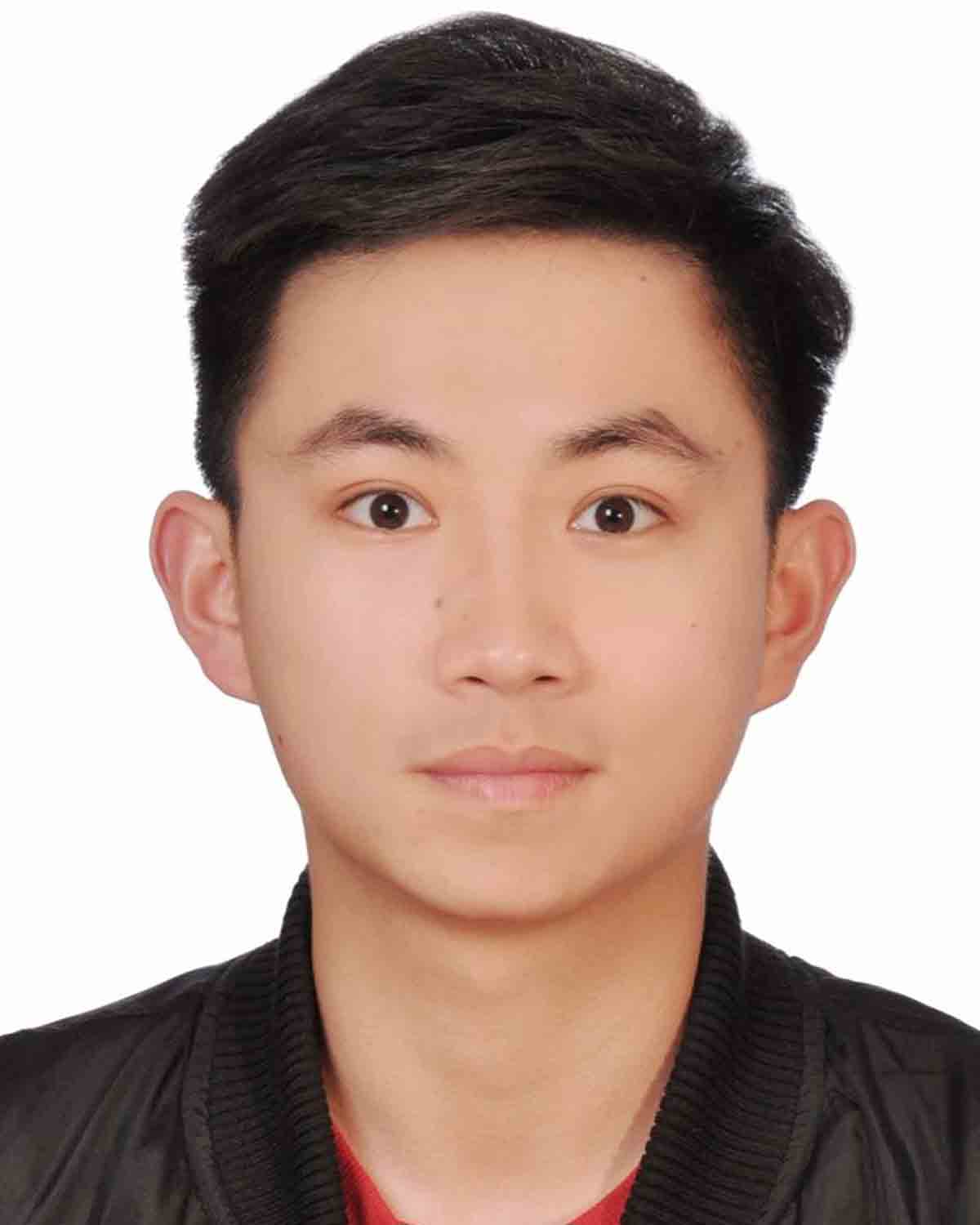}}]
{Zhengxu Yu} is now a doctoral candidate at Zhejiang University, and he is currently a research intern at Alibaba DAMO Academy. His research interests include large scale machine learning and computer vision.
\end{IEEEbiography}

\begin{IEEEbiography}
[{\includegraphics[width=1in,height=1.25in]{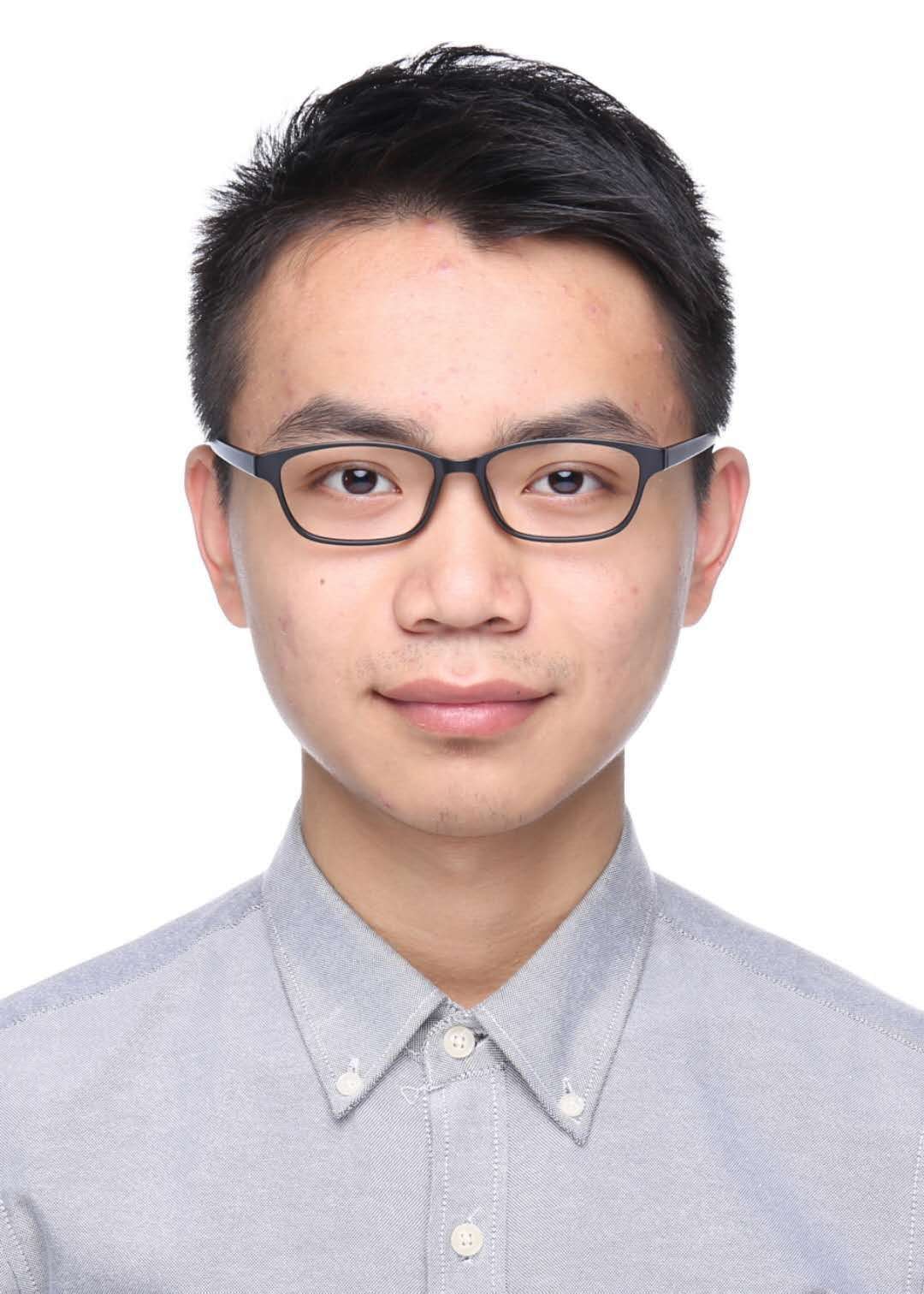}}]
{Dong Shen} received a B.S. degree in computer science from Zhejiang University, China, in 2018. He is currently a master student in computer science at Zhejiang University. His research interests include machine learning and computer vision.
\end{IEEEbiography}

\begin{IEEEbiography}
[{\includegraphics[width=1in,height=1.25in]{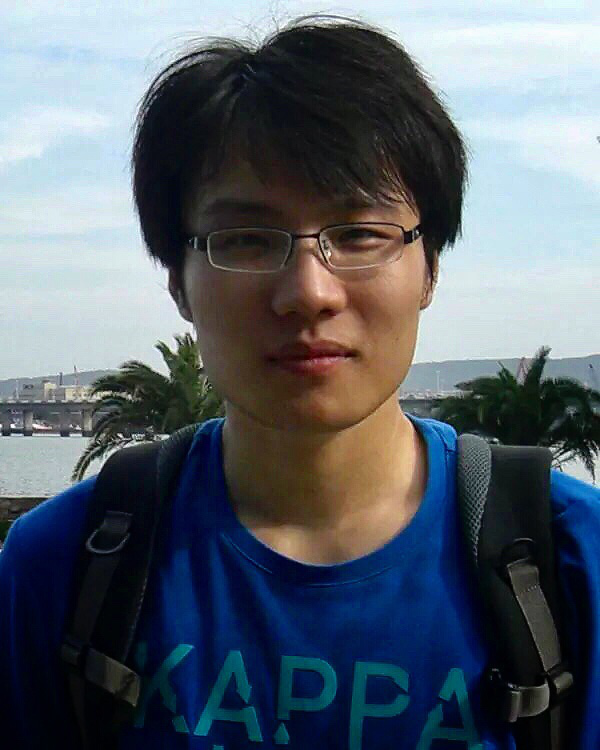}}]
{Zhongming Jin} is now a staff algorithm engineer at Alibaba DAMO Academy. Previously, he was a researcher at Baidu Research. He received his Ph.D. degree from Zhejiang University in Mar. 2015. His research interests include large scale machine learning and computer vision.
\end{IEEEbiography}

\begin{IEEEbiography}
[{\includegraphics[width=1in,height=1.25in]{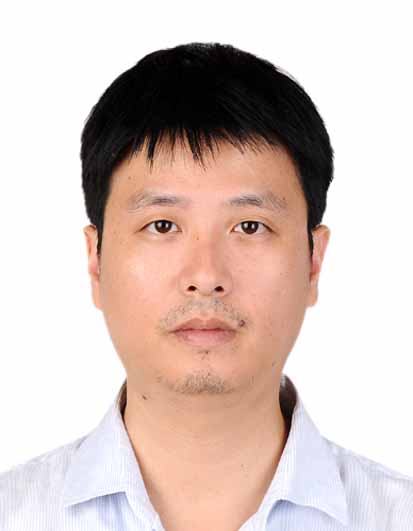}}]
{Jianqiang Huang} is a director of Alibaba DAMO Academy. He received the second prize of National Science and Technology Progress Award in 2010. His research interests focus on visual intelligence in the city brain project of Alibaba.
\end{IEEEbiography}

\begin{IEEEbiography}
[{\includegraphics[width=1in,height=1.25in]{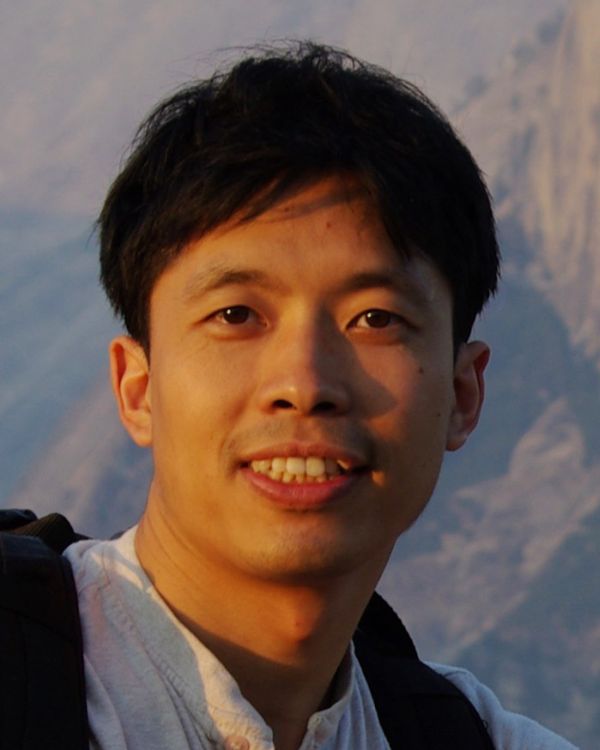}}]
{Deng Cai} is a Professor in the State Key Lab of CAD\&CG, College of Computer Science at Zhejiang University, China. He received a Ph.D. degree in computer science from the University of Illinois at Urbana Champaign in 2009. His research interests include machine learning, data mining and information retrieval.
\end{IEEEbiography}

\begin{IEEEbiography}
[{\includegraphics[width=1in,height=1.25in]{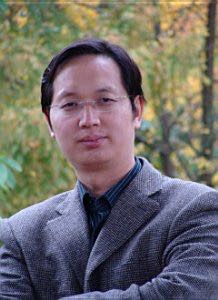}}]
{Xian-Sheng Hua} (F'16) received the B.S. and Ph.D. degrees in applied mathematics from Peking University, Beijing, in 1996 and 2001, respectively. In 2001, he joined Microsoft Research Asia as a Researcher
and has been a Senior Researcher of Microsoft Research
Redmond since 2013.
He became a Researcher and the Senior Director of the Alibaba Group in 2015.
He has authored or co-authored over 250 research papers and has filed over 90 patents. His research interests have been in the areas of multimedia search, advertising, understanding, and mining, and pattern
recognition and machine learning. 
He was honored as one of the recipients of MIT35.
He served as a Program Co-Chair for the IEEE ICME 2013, the ACM Multimedia 2012, and the IEEE ICME 2012, and on the Technical Directions Board of the IEEE Signal Processing Society. He is an ACM Distinguished Scientist and IEEE Fellow. 
\end{IEEEbiography}
\vfill

\end{document}